\documentclass[a4paper,12pt,oneside]{article}
\usepackage{amssymb,amsfonts,amsmath,mathtext,cite,enumerate,float,}
\usepackage[dvips]{graphicx}

\makeatletter
\renewcommand{\@biblabel}[1]{#1.} 
\makeatother

\usepackage{geometry}
\begin{document}
\author{M.V. Konnik, E.A. Manykin, S.N. Starikov}
\title{INCREASING LINEAR DYNAMIC RANGE OF COMMERCIAL DIGITAL PHOTOCAMERA USED IN IMAGING SYSTEMS WITH OPTICAL CODING}
\date{Moscow Engineering Physics Institute (State University)}
\maketitle

\begin{abstract}
Methods of increasing linear optical dynamic range of commercial photocamera for optical-digital imaging systems are described. Use of such methods allows to use commercial photocameras for optical measurements. Experimental results are reported.
\end{abstract}

\section{Introduction}
Increased interest to hybrid optical-digital systems based on ``wavefront coding''~\cite{wavefrontcodingMems} and ``pupil engineering''~\cite{grachtPupil} principles is observed recently. Such systems allow to create devices, which combine highly paralleled optical processing and flexibility of digital image processing techniques. This conducing improvement of mass and dimensions of such systems and allows to create devices with unachievable characteristics for pure optics.

Imaging scheme of such kind of hybrid systems is changed by inserting synthesized diffraction optical element - kinoform. This enables to perform optical convolution and register optically convolved image by digital photosensor. It's notable that all information about input scene is contented in greyscale levels of the coded image. Non-linear image processing like gamma-correction, image interpolation, and colour scaling can corrupt coded image and bring to difficulties in digital deconvolution. That's why linear registration of images by photosensor is important. 

In this paper linearization~\cite{davecoffin,konnikLinearizationHoloexpoEng} and spatially varying pixel exposures~\cite{mitsunagaSVErecover} techniques are applied for increasing of full and linear dynamic ranges of the images obtained from commercial photocameras. Using special software allows to exploit linearity of camera's sensor and apply inexpensive cameras in optical-digital systems were linearity of signal is highly important. 

This paper is organised as follows. Camera's radiometric function is provided both for DCRAW converter and Canon's conventional converter in subsection~\ref{sec:radiometric}, dynamic range estimation  is provided in subsection~\ref{sec:dynamicrange}. Dark noise and light-depended noise estimation is presented in subsections~\ref{sec:darknoise} and \ref{sec:lightnoise} respectively. Increasing of camera's images dynamic range using Spatially Varying pixel Exposures technique is discussed in section~\ref{sec:SVE}.

\section{Commercial photocamera as measuring device}
In this section the linearization procedure and measuring characteristics of the commercial digital camera Canon EOS 400D are described. The radiometric function was measured in order to estimate camera's sensor linearity and saturation level. Linear and full dynamic range, black level offset (BLO), temporal and spatial noise were evaluated as well. All RAW images were processed by the DCRAW~\cite{davecoffin}  converter using ``document mode'' without colour scaling and interpolation. For viewing and analysing of such images it is suitable to use the \textit{NIP2}~\cite{ecs12371,vipsspie} open source image analyser.

\subsection{Camera's radiometric functions obtained with different converters}\label{sec:radiometric}
\begin{figure}[H]
	\begin{minipage}[h]{0.49\linewidth}
	\center{\includegraphics[width=1\linewidth]{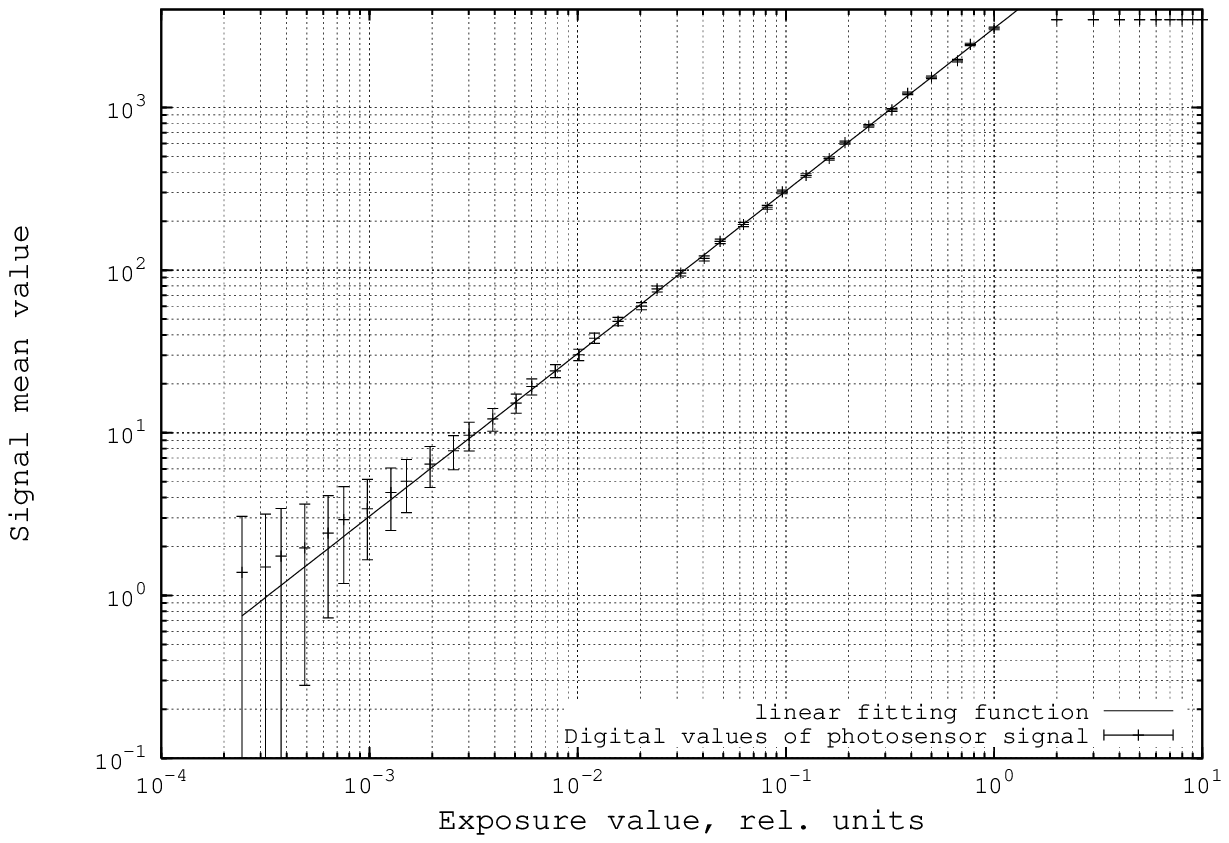} \\ a)}
	\end{minipage}
\hfill
	\begin{minipage}[h]{0.49\linewidth}
	\center{\includegraphics[width=1\linewidth]{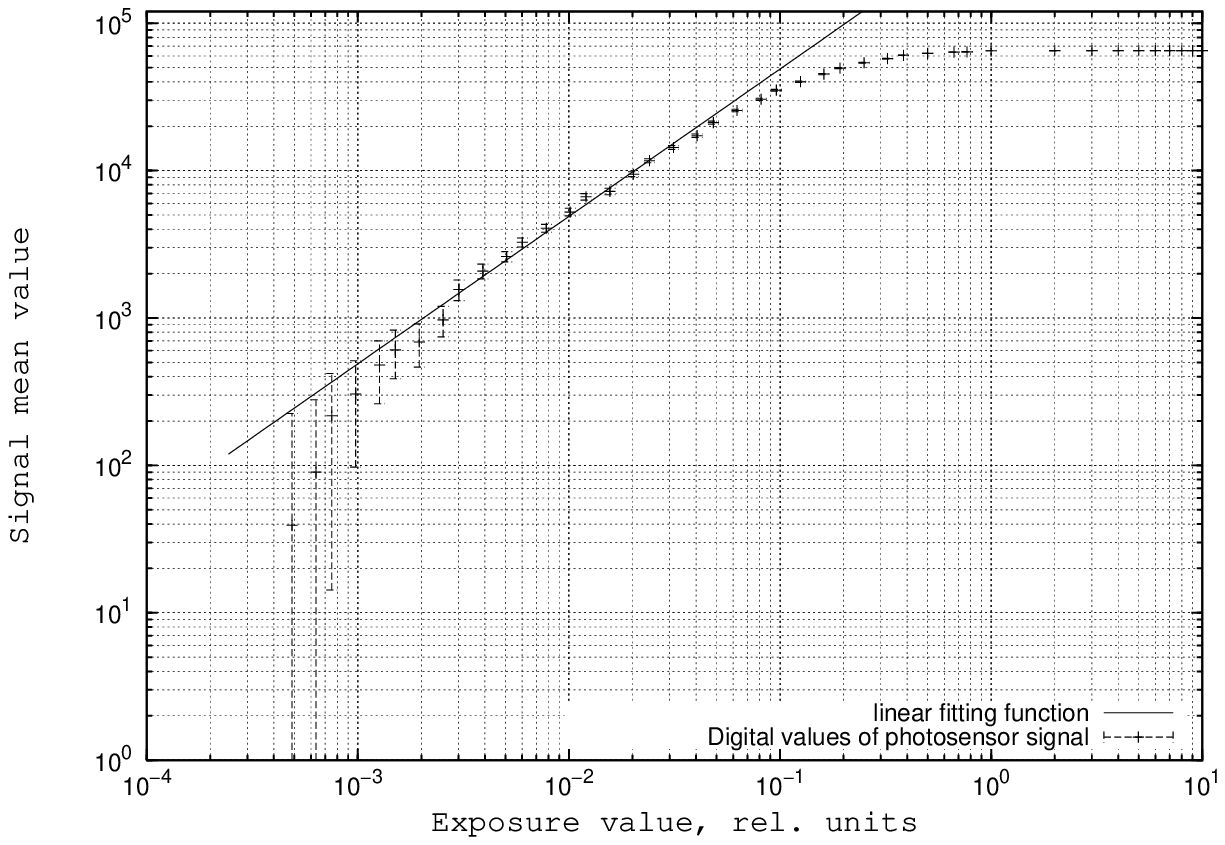}   \\ b)}
	\end{minipage}
	\caption{Signals' mean value in DN versus relative exposure: a) for DCRAW processed images, b) for images processed by conventional Canon RAW-converter.}
	\label{ris:1RAWMEANtoSaturateLineApproxcomparing}
\end{figure}

For the radiometric function to obtain, pictures of the flat field scene in the light of white LEDs were taken. Light was passed through ground glasses to remove flat field's non-uniformity, images were taken in a laboratory without any lighting and reflections from the surrounding surfaces. Pictures were taken with the exposure time from 1/4000 to 10 seconds,  4 images were taken for each exposure and averaged.  ISO setting was the smallest available for Canon EOS 400D camera (ISO 100). Pictures were processed by DCRAW converter in ``document mode'' by command \textit{dcraw -4 -T -D -v filename.cr2} to produce pure RAW 12-bit images. Also images were processed by conventional Canon converter and obtained colour images were decomposed to three greyscale images. Dependencies presented in Fig.~\ref{ris:1RAWMEANtoSaturateLineApproxcomparing} are for blue-filter covered pixels; curves for all colour-filtered pixels differs only by a constant shift along the exposure axis.

A 64 by 64 pixel area from the centre of the image was used for the analysis for every colour-filter type. During processing of obtained data, black level offset (BLO) of 256\;DN for DCRAW processed images was disclosed and then removed. Canon converter subtracts BLO internally. Mean and standard deviation of pixels values were obtained and results are presented in Fig.~\ref{ris:1RAWMEANtoSaturateLineApproxcomparing}a for DCRAW converter and in Fig.~\ref{ris:1RAWMEANtoSaturateLineApproxcomparing}b for conventional Canon converter. Saturation level for the DCRAW converted data is equal 3726\;DN, and for Canon converted data saturation level is equal to 65535\;DN. One can see that DCRAW processed data for radiometric function is linear up to saturation level.

%
%
%

\subsection{Camera's dynamic range estimation}\label{sec:dynamicrange}
\begin{figure}[H]
	\begin{minipage}{0.49\linewidth}
		\center{\includegraphics[width=1\linewidth]{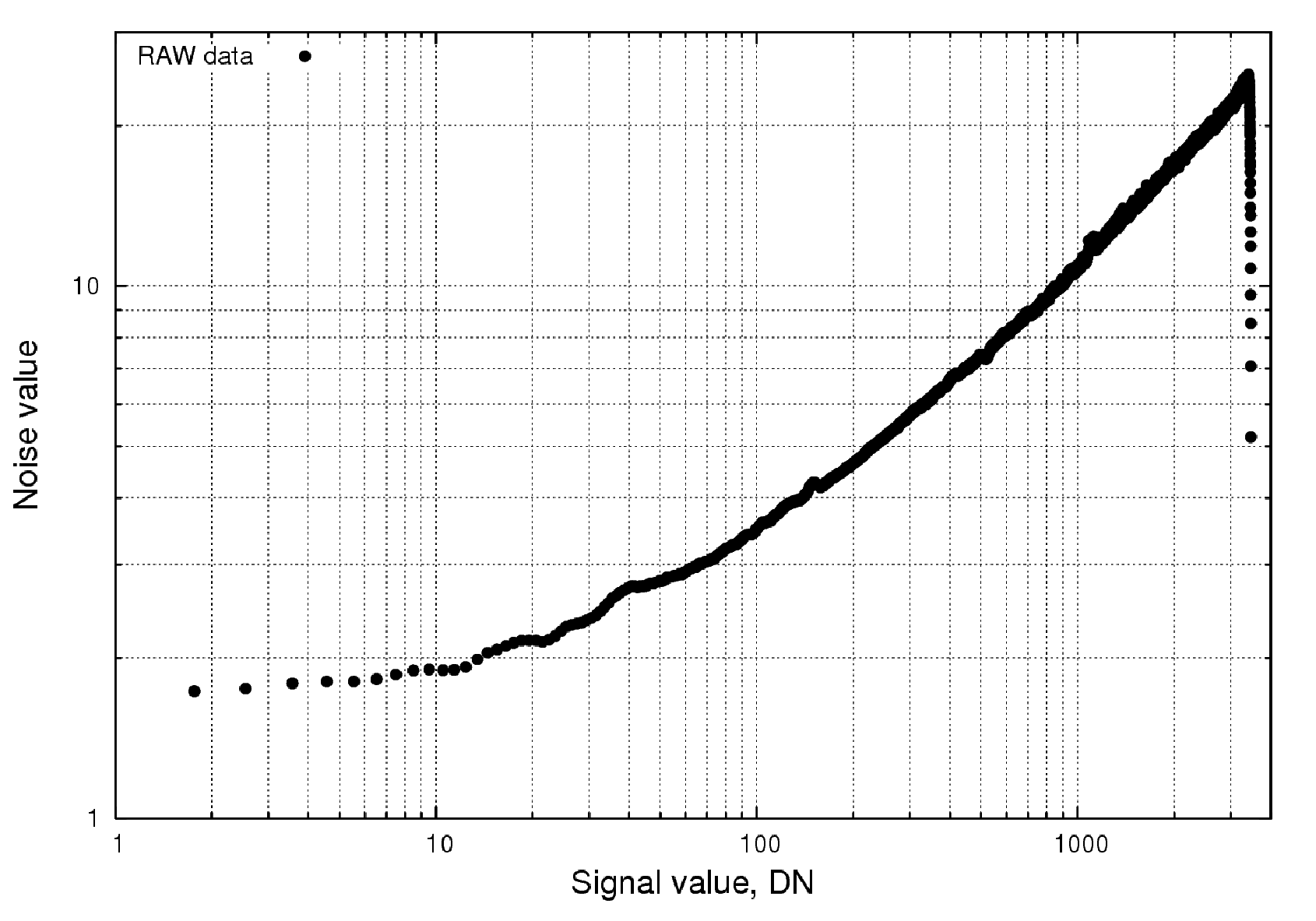}} \\ a)
	\end{minipage}
\hfill
	\begin{minipage}{0.49\linewidth}
	\center{\includegraphics[width=1\linewidth]{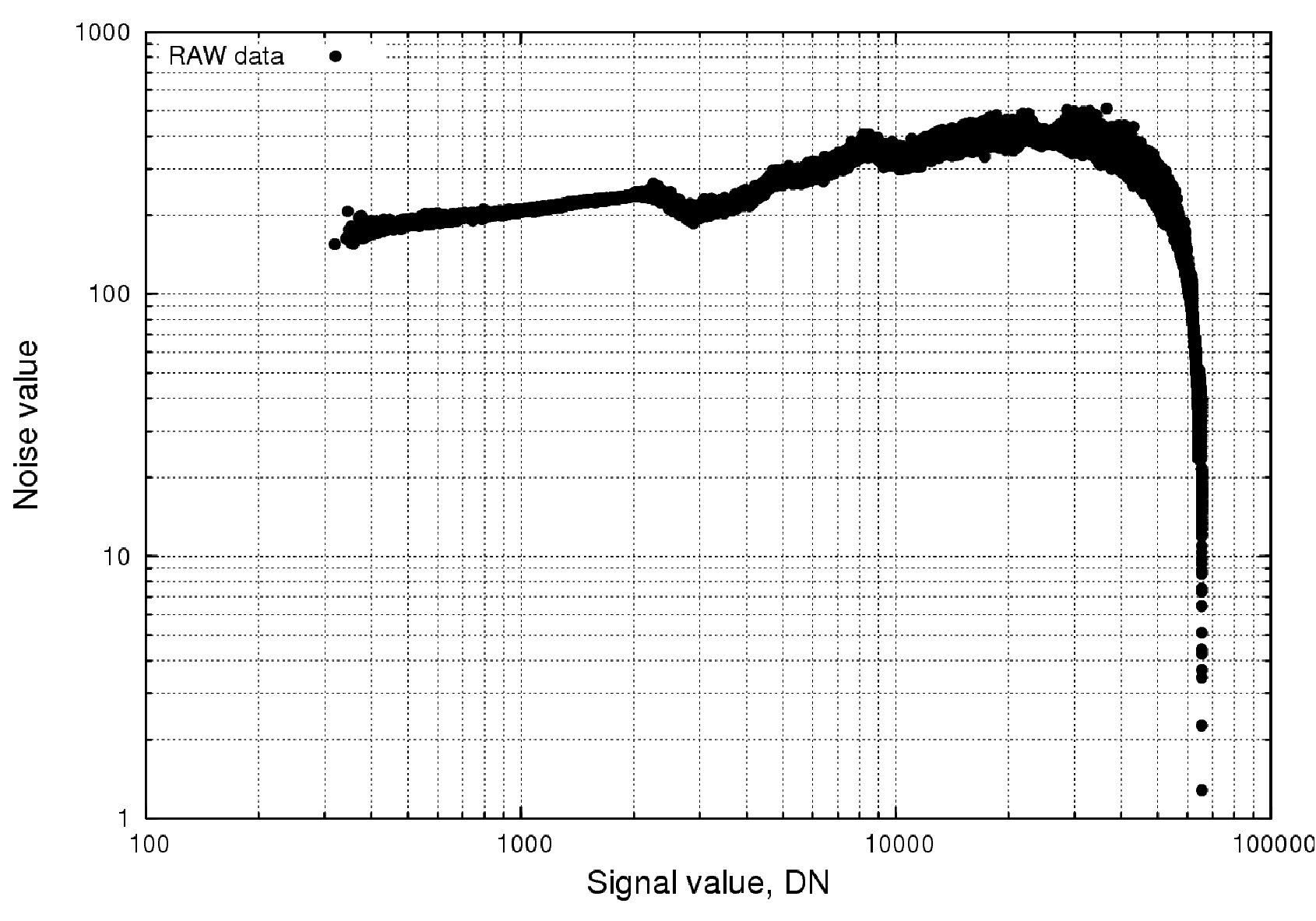}} \\ b)
	\end{minipage}
	\caption{Noise versus signal (both in DN): a) for DCRAW converted data, b) for conventional Canon converter.}
	\label{ris:NoiseVSSignallogmean64Bluestd64BlueSignalstep1Bias256SIGNALNOISEmeasurementresult}
\end{figure}

For estimation of the dynamic range's beginning, the minimal SNR was considered as 2. The minimal detectable signal corresponds to SNR=2 is equal 4\;DN (see Fig.~\ref{ris:NoiseVSSignallogmean64Bluestd64BlueSignalstep1Bias256SIGNALNOISEmeasurementresult}a). It is possible to find out relative exposure value that corresponds to the minimal signal: it equals to $1.3\cdot10^{-3}$ rel. units, as seen from  Fig.~\ref{ris:1RAWMEANtoSaturateLineApproxcomparing}a.

A linear function was fitted in experimental data of the radiometric function; then signal's value corresponding to the end of a linear dynamic range can be estimated as 3066\;DN (see Fig.~\ref{ris:1RAWMEANtoSaturateLineApproxcomparing}a) with relative exposure time 1.0 rel. units for this point. Therefore the linear dynamic range with DCRAW data processing is 58\;dB. To estimate full dynamic range is necessary to measure of the maximum saturation signal and minimal detectable signal. Using the same calculation procedure described above and SNR as 2, minimal distinguishable signal remains 4\;DN with relative exposure time $1.3\cdot10^{-3}$ rel. units. Maximum detectable signal is 3438\;DN, hence relative exposure value for this point is 1.12 rel. units. Therefore full dynamic range can be estimated as 59\;dB.

The same calculations were provided for data converted by Canon conventional converter. Using obtained radiometric function presented in Fig.~\ref{ris:1RAWMEANtoSaturateLineApproxcomparing}b and taking into account Fig.~\ref{ris:NoiseVSSignallogmean64Bluestd64BlueSignalstep1Bias256SIGNALNOISEmeasurementresult}b, it is possible to estimate linear and full dynamic ranges for non-linearized data. As above, minimal signal to noise ratio is 2; thus one can estimate minimum signal value as 400\;DN and corresponding relative exposure value is $8.2\cdot10^{-4}$ rel. units. Maximum value of linear signal is estimated by fitting linear function to experimental data (see  Fig.~\ref{ris:1RAWMEANtoSaturateLineApproxcomparing}b) and it equals 24454\;DN. Corresponding relative exposure value for such signal is $5.0\cdot10^{-2}$ rel. units. Thus linear dynamic range is considered to be 36\;dB.

As follows from the obtained data, DCRAW converter allowed to use linear dynamic of  58~dB from 59~dB of the full dynamic range. By contrast, conventional Canon converter can provide only 36~dB of the linear dynamic range.

It is significantly to note that obtained results are only an \textit{approximation} of the commercial camera's photo sensor characteristics because of sufficient dispersion of noise characteristics between cameras of same model and presence of the on-chip circuitry. It is discussed in subsection~\ref{sec:discussion}.

\subsection{Dark noise estimation}\label{sec:darknoise}
Noise components can be classified in different ways. One type of classifications divides the noise components into random (temporal) and pattern (spatial) components. Random components include photon shot noise, dark current shot noise, reset noise, and thermal noise. Pattern components are amplifier gain non-uniformity, PRNU, dark current non-uniformity, and column amplification offset. In the commercial cameras it is difficult to distinguish fine structure of noise components because of on-chip noise-cancelling circuitry. Hence only general noise components description and estimation is provided such as dark and light-depended noise temporal and spatial components.

\subsubsection{Spatial dark noise}
Spatial dark noise can be referred as pixel-to-pixel variation of the dark signal. In the modern commercial CMOS cameras it is difficult to measure spatial dark noise because of presence of the noise-cancelling \textit{on-chip circuitry}~\cite{canonfullframeCMOSwhitepaper}; more detailed discussion of this question is provided in Section~\ref{sec:discussion}. 

In order to estimate the spatial dark noise, there were taken 64 dark frames with capped camera's objective. The ISO speed was 100 and exposure time was 1/32 sec. Then mean value and variance along the averaged dark frame were calculated. 
%
According to carried out measuring, mean value of the averaged dark frame is 256.0 DN and its standard deviation is $\sigma_{dark.spat} \approx 0.4$ DN. 
%
%
%
We are emphasizing that such low spatial dark noise is due to the on-chip circuitry noise reduction of the Canon's digital camera.

\subsubsection{Temporal dark noise}\label{sec:darktempnoise}
Temporal component of the dark noise was also estimated. For such purpose there were taken 64 dark frames. Then arrays of pixels were averaged and the RMS noise of each pixel was calculated. After such procedure two another arrays are created: the array of pixel's mean values $A_{mean}$ and the array of pixel's standard deviations $A_{std}$ (and consequently the array of pixel's variations $A_{var}$). This procedure is analogous to the PixeLink's method~\cite{pixelLinkCameraParams}.

To estimate the temporal dark noise quantitatively, the average variation of the $A_{var}$ need to be calculated and square root is taken. Consequently, the temporal dark noise can be evaluated as follows:

\begin{equation}
\sigma_{dark.temp} = \sqrt{\frac{1}{MN} \sum_{i,j} \sigma^2_{dark.temp.ij} },
\end{equation}
where M and N is the heigh and width of the dark frame, respectively. To measure deviation of dark temporal noise between pixels, one should calculate the standard deviation from $A_{std}$.

For the digital camera used in this work there were averaged 64 dark frames (ISO speed was 100, exposure time 1/32 sec.). Thus the $\sigma_{dark.temp} \approx 1.6$~DN, and uncertainness of the temporal dark noise is 0.2~DN. 

\subsection{Light-depended noise estimation}\label{sec:lightnoise}
The light-depended noise was evaluated as well. There were taken and averaged images of the flat-field scene. The lighting used was matrix of red, green, and blue LEDs driven with DC current. ISO setting was ISO 100, the smallest available in the camera. Objective was removed in order to achieve flat-field homogeneity. A 1024 by 1024 pixel area from the centre of the image was used for the analysis.

\subsubsection{Spatial light-depended noise}
As a measure of the spatial light-depended noise, photo-response non-uniformity (PRNU) is commonly used~\cite{noiseCharacterizationCameras,pixelLinkCameraParams}. PRNU is the standard deviation of the flat-field image with subtracted an averaged dark frame.

There were taken and averaged 64 pictures of the flat-field scene. Then averaged dark frame was subtracted from averaged picture of the flat-field. Obtained picture was decomposed on three images: pixels corresponded to red colour filters were stored in $B_r$ array, pixels corresponded to first green colour filter were stored in $B_g$ array, and pixels corresponded to blue colour filters were stored in $B_b$ array. Then for each array there were calculated a standard deviation divided by the frame mean value. Thus PRNU for each colour component was evaluated as follows:
\begin{equation}
	PRNU = \frac{\sigma_{light.spat}} {FrameMean} \%
\end{equation}

According to our measures, PRNU can be estimated as $PRNU \leq 0.5 \%$ ($\sigma_{light.spat} \approx 12$~DN and mean value of the signal is 2600). It is needed to mention that such PRNU is the \textit{residual PRNU} after on-chip circuitry noise-reduction that is performed by electronics of the digital camera and can not be turned off.

\subsubsection{Temporal light-depended noise}
Temporal light-depended noise is an uncertainness of light's measuring by each pixel, hence the calculation procedure is analogous to the procedure for temporal dark noise  evaluation (see Subsection~\ref{sec:darktempnoise}). 

The RMS values for each pixel of the flat-field scene's image was calculated, forming two another arrays: the array of pixel's mean values $A_{mean}$ and the array of pixel's variations $A_{var}$. Then obtained array was decomposed accordingly to the light components R, G, and B, same as for PRNU estimation. 

Hence temporal light-depended noise was evaluated for each colour channel separately:
\begin{equation}
\sigma_{light.temp} = \sqrt{\frac{1}{MN} \sum_{i,j} \sigma^2_{light.temp.ij} }
\end{equation}


Thus it can be summarized that temporal light-depended noise for each colour channel can be estimated as $\sigma_{light.temp} \approx 14$ DN.

\subsection{Results discussion}\label{sec:discussion}
It is important to note that presented results are only an \textit{approximation} of the commercial camera's photo sensor characteristics. This is because of on-chip noise reduction circuitry and sufficient dispersion of noise characteristics between cameras of same model. 

In scientific grade technical cameras there are calibration methods controlled by user. By contrast, in commercial cameras are used their own proprietary noise-cancelling technologies that reduces both dark and light-depended noise. As it mentioned in~\cite{canonfullframeCMOSwhitepaper}, Canon developed on-chip technology to reduce fixed-pattern noise based on Correlated Double Sampling~\cite{correlateddoublesampling}. First, only the noise is read. Next, it is read in combination with the light signal. When the noise component is subtracted from the combined signal, the fixed-pattern noise can be eliminated.

Moreover, light-depended noise also is been suppressed~\cite{canonfullframeCMOSwhitepaper} by on-chip circuitry. Such method is called \textit{complete electronic charge transfer}, or complete charge transfer technology. Canon designed the photodiode and the signal reader independently to ensure that the sensor resets the photodiodes that store electrical charges. By first transferring the residual discharge - light and noise signals - left in a photodiode to the corresponding signal reader, the sensor resets the diode while reading and holding the initial noise data. After the optical signal and noise data have been read together, the initial noise data is used to remove the remaining noise from the photodiode and suppress random noise. Thus only an \textit{estimation} of the commercial camera's noise characteristics is possible.

\section{Further increasing of the camera's dynamic range}\label{sec:SVE}
\begin{figure}[H]
\begin{center}
	\includegraphics[width=0.5\linewidth]{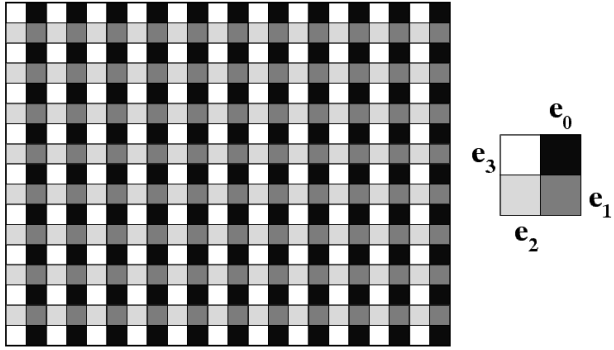}
	\caption{The dynamic range of an image detector can be improved by assigning different exposures to pixels.}
	\label{ris:SVEmatrixpatter}
\end{center}
\end{figure}

Besides linearization, several techniques such as spatially varying pixel exposures~\cite{mitsunagaSVErecover} (SVE) and Assorted Pixels~\cite{sveassorted} can be applied for further increase of camera's dynamic range. The key idea of such techniques is to use data from colour filters on photosensor to estimate true value of oversaturated pixels (Fig.\ref{ris:SVEmatrixpatter}).

To reconstruct a linear HDR image from the oversaturated one, several steps should be performed. Such steps are the calibration, construction of the SVE-image, and linearization of the constructed image. First, the camera's response function to the desirable lightsource is obtained. During this calibration process, the camera's response function to the lightsource as well as correction coefficients for SVE-images linearization are calculated. Secondly, an oversaturated image is analysed and saturated pixels are replaced using information from the neighbour pixels. Such constructed SVE-image is characterised by a $\gamma$-like non-linearity. Finally, constructed SVE-image is linearized using correction coefficients, which were obtained on the first (calibration) step. Obtained linear HDR image is characterised by a broad dynamic range and a linear response to the registered light. 

It is necessary to estimate the amount of quantization levels that can be obtained using SVE technique. If total number quantization levels is $q$ and the number of different exposures in the pattern is $K$, then total of $q\cdot K$ quantization levels lie within the range of measurable radiance values. Only the quantization levels contributed by the highest exposure within any given overlap region. Total number of unique quantization levels according to~\cite{mitsunagaSVErecover} can be determined as:

\begin{equation}\label{eq:fullSVEhalftones}
Q = q + \sum \limits_{k=1}^{K} { R 
\Bigl(
(q-1) - (q-1) \frac{e_{k+1}}{e_{k}}
\Bigr)
}
\end{equation}
where R(x) rounds-off x to the closest integer. Hence using Eq.~\ref{eq:fullSVEhalftones} one can estimate quantization levels for Canon EOS 400D commercial photo camera used in this work. Using DCRAW converter, it is possible to obtain 3066 linear quantization levels for each pixel.  Assume that $e_1$ is transmittance of red light, $e_2$ is transmittance of green light, and $e_3$ of blue light. Experimentally for $\lambda = 0.63 \mu$m He-Ne laser radiation were obtained that $e_2/e_1 = 0.2$ and $e_3/e_2 = 0.45$. Substituting this into Eq.~\ref{eq:fullSVEhalftones}, $Q \approx 7200$ quantization levels can be obtained, which is a considerable improvement over q = 3066 for the same image detector.

\section{Conclusion}
The use of linearized RAW data from the commercial photo camera and its adoption for measurements is presented. Using DCRAW converter with proper parameters, it is possible to exploit linearity of camera's photo sensor response to registered light and use the commercial camera as a measuring device. 

For Canon EOS 400D digital camera used in this work with DCRAW converted data were obtained the following characteristics: linear dynamic range is 58~dB and full dynamic range 59~dB with 12bit ADC.  Spatial dark noise is $\sigma_{dark.spat} \approx 0.4$~DN, and bias is 256.0 DN. Temporal dark noise can be estimated as $\sigma_{dark.temp} \approx 1.6$~DN and its uncertainness is 0.2~DN. Photo Response Non-Uniformity(PRNU) is less than 0.5\% ($\sigma_{light.spat} \approx 12$~DN, mean value of the signal is 2600) for all colour channels. Temporal light-depended noise can be estimated as $\sigma_{light.temp} \approx 14$~DN.

For further increasing of camera's dynamic range Spatially Varying pixel Exposures technique can be used. As it shown above, in a quasimonochromatic light it is possible to receive around 7200 quantization levels of the input signal, which is a considerable improvement over 3066 for the same image detector.

Obtained results are an approximation of the camera's photo sensor characteristics because of sufficient dispersion of noise characteristics between cameras of same model and presence of the noise-reduction on-chip circuitry. 

Application of described linearization method allows to increase the dynamic range of images produced by commercial digital photocamera. From carried out experiments it follows that inexpensive commercial photo cameras can be used in optical-digital imaging systems.

\end{document}